\documentclass{isprs} 
\usepackage{subfigure}
\usepackage{setspace}
\usepackage{geometry}
\usepackage{epstopdf}
\usepackage{adjustbox}
\usepackage{amssymb}
\usepackage{tikz}
\usepackage[labelsep=period]{caption}  
\usepackage[british]{babel} 
\usepackage[hang]{footmisc}
\usepackage{hyperref}
\hypersetup{
    colorlinks=true,
    allcolors=black,
    linkcolor=black,
    urlcolor=blue,
    citecolor=black
}
\begin{document}

\title{SYNTHETIC DATA GENERATION PIPELINE FOR GEOMETRIC DEEP LEARNING IN ARCHITECTURE
}
\version{}
\author{S. Fedorova\textsuperscript{1}\thanks{Corresponding author}\ , A.Tono\textsuperscript{2}\thanks{denotes equal contribution} ,
M. S. Nigam\textsuperscript{3}\footnotemark[2] , 
J. Zhang\textsuperscript{4}, 
A. Ahmadnia\textsuperscript{1} , 
C. Bolognesi\textsuperscript{1} ,
D. Michels\textsuperscript{4}}
\address{
\textsuperscript{1}Politecnico di Milano, Milan, Italy - (stanislava.fedorova, amirhossein.ahmadnia)@mail.polimi.it, cecilia.bolognesi@polimi.it \\
\textsuperscript{2}CDI, Computational Design Institute, San Francisco - alberto.tono@cd.institute \\
\textsuperscript{3}IIIT Hyderabad, India -  meher.shashwat@students.iiit.ac.in \\
\textsuperscript{4}KAUST, Saudi Arabia - jiayao.zhang@kaust.edu.sa
}
\icwg{}   

\abstract{
With the growing interest in deep learning algorithms and computational design in the architectural field, the need for large, accessible and diverse architectural datasets increases. Due to the complexity of such 3D datasets, the most widespread techniques of 3D scanning and manual building modeling are very time-consuming, which does not allow to have a sufficiently large open-source dataset. 
We decided to tackle this problem by constructing a field-specific synthetic data generation pipeline that generates an arbitrary amount of 3D data along with the associated 2D and 3D annotations. The variety of annotations, the flexibility to customize the generated building and dataset parameters make this framework suitable for multiple deep dearning tasks, including geometric deep learning that requires direct 3D supervision. Creating our building data generation pipeline we leveraged the experts' architectural knowledge in order to construct a framework that would be modular, extendable and would provide a sufficient amount of class-balanced data samples. Moreover, we purposefully involve the researcher in the dataset customization allowing the introduction of additional building components, material textures, building classes, number and type of annotations as well as the number of views per 3D model sample. In this way, the framework would satisfy different research requirements and would be adaptable to a large variety of tasks. All code and data is made publicly available: \href{https://cdinstitute.github.io/Building-Dataset-Generator/}{cdinstitute.github.io/Building-Dataset-Generator}.
}
\maketitle

\section{Introduction}
The recent advances in CAD software showed the advantages of using 3D models to the specialists working in various fields. The advantages include the possibility of a 3D visualization of a project, inclusion of all the relevant metadata related to it, a relatively quick transformation to the desired representation, and the modifications that are reflected on all the stages and views. It is especially relevant to the field of architecture and urban planning that involves a lot of different data required to take into consideration as well as the spatial representation of a project. At the same time, the time and resources required for a full 3D model creation for a building project are one of the obstacles that prevent some of the specialists of the field from using 3D modeling in their pipelines.
\par A potential solution to this issue could be leveraging the Geometric Deep Learning techniques for the generation of 3D models from a given input. This input could potentially be represented in various ways, for instance, as a set of constraints determining the project or an image representation. In the case of an architectural project, it would be quite time-consuming to determine a set of rules due to the complexity of the subject. The use of the images would take advantage of the ability of Deep Neural Networks to learn visual features in a self-supervised manner.
\par One of the challenges the researchers face when dealing with this question is availability of annotated 3D datasets. The complexity of the 3D model creation process, a variety of software and approaches to building modeling, differences in the modeling quality, and the lack of access for the completed and finished 3D models make it difficult for the neural networks to learn successfully. At this point, to our knowledge, there are no existing architectural datasets at the building scale with a sufficient amount of samples to perform Geometric Deep Learning. Due to these reasons we provide the architectural community with a data generation pipeline that automatically creates a synthetic building dataset suitable for various Deep Learning tasks. Due to the specificity of our contribution and the lack of the datasets on this architectural scale, there are no benchmarks or downstream metrics existing so far.
\par
The establishment of a 3D dataset of building envelopes with relevant information and development of a 3D reconstruction framework could benefit a manifold of different industries such as robot navigation, drone delivery, retails, urban planning, AR/ VR gaming experiences.

\section{Related Works}
\textbf{3D Datasets}
The complexity of the urban planning and architecture fields create a variety of Deep Learning applications applicable to them. The drawback of this aspect is that different tasks require different input and ground truth data not only in terms of format but also in terms of internal requirements within one format. For instance, generative design of a building facade would require a semantic segmentation annotation, while 3D reconstruction would require a 2D representation of a building as well as a 3D representation, the format of which could vary based on the researcher's approach (mesh, voxels, point cloud). The intricacy of the subject of interest makes the dataset creation time-consuming, so while there are quite a few 3D datasets \cite{pix3d} \cite{chang2015shapenet} \cite{modelnet} \cite{mcb} \cite{pascal3d} \cite{objectnet3d} \cite{partnet} \cite{abc} \cite{ikea} (Table~\ref{tab:dataset_analysis}), not many of them are related to architecture. The format issue makes this scarce number even smaller when applied to a specific problem, such as facade reconstruction, 3D classification or semantic segmentation of the building parts.
\newline Urban 3D datasets are the most common ones in the field. This is due to the availability of data at the city level (shapefiles), open-source geospatial software and the relative ease of production, as most of these datasets contain Level of Detail 1 (LoD1) models \cite{sum2021}. Level of Detail(LoD) is an important characteristic of the architectural datasets that allows to specify the amount of detail and generalization present in the 3D model \cite{BILJECKI201625}. LoD1 refers to a building envelope without any additional details while LoD4 indicates a detailed building model with an internal structure.
\par The work that that aims to solve the same problem is 3DCityDB \cite{3dcitydb}, which contains the 3D representations of the real cities. The advantage their system has, is the use of the real-world data and texturing that gives more precise and realistic representations of buildings. The drawbacks of their framework include the limited amount of samples (as for the time of writing the database mentions the cities of Berlin and New York), the lack of meaningful information about the structures, absence of the semantic segmentation related to the parts of the buildings and the simplified models of the buildings (extruded polygons) that do not include finer details. The models in 3DCityDB are given in the urban context which could be seen as an advantage or disadvantage based on the task; moreover, there are no image annotations or separate 3D files the researchers could use directly for Deep Learning purposes.
\par Another approach that is similar to ours is Random3DCity\\ \cite{random3dcity}, that exploits procedural generation in order to create a variety of building forms. However, the authors aim to get a simplified building representation without the use of the real-world texturing which makes it inapplicable to single image to 3D reconstruction problem as we approach it. Moreover, the tool is intended to be used in CityGML format \cite{cityGML} that imposes certain limitations on the researchers. The main advantage of this dataset is the possibility to decide the LoD level up to LoD4 which includes the internal building structure and modularity that allows a relatively easy scaling of the dataset and its variations.
\par The dataset that is closest to ours is Structured3D \cite{structured3d} that involves synthetic 3D models and image annotations for the architectural interior space. The annotations of this dataset include segmentation, depth and rendered images as well as the objects' structure and several interior configurations. There are a few more synthetic interior 3D scene datasets such as SceneNet \cite{scenenet} or InteriorNet \cite{InteriorNet18}  (does not include 3D models). Unfortunately, all of them tackle the problem at a different architectural scale concentrating on one interior space while our task requires building exterior models with the information related to the structure.
\par Another disadvantage of the mentioned architectural datasets is related to their construction process, as the building objects they contain are not parametric, even the ones containing synthetic data. This aspect puts some limitations on the researchers using these datasets. We address this issue in our solution by providing a parametric building generation using programmatic procedural approach.

\section{Dataset Generation Pipeline}
\subsection{Existing Datasets}

\begin{table}[!ht]
\begin{center}
\begin{adjustbox}{max width=\columnwidth}
		\begin{tabular}{|l|c|c|c|c|}
		\hline
			\textbf{Dataset}&\textbf{Classes}&\textbf{Samples}&\textbf{Images}&\textbf{Std-dev}\\
			\hline
			 IKEA&7&219&759&16.4 \\
			 ShapeNetCore & 55 & 51,300   &   307,800 &   1589.7 \\
			 ShapeNetSem & 970 & 9,033  &    72,000 &   343.3 \\
			 PartNet & 24 & 26,600 & NA & 322.4 \\
			 ObjectNet3D & 100 & 44,147 & 90,127 & 42.0 \\
			 Pascal3D+ & 12 & 79 & 30,899 & 1.9 \\
			 ModelNet10 & 10 & 4,899 & NA & 243.9 \\
			 ModelNet40 & 40 & 12,311 & NA & 215.7  \\
			 MCB & 68 & 58,700 & +300,000 & 1040.1 \\
			 ABC & NA & 1,000,000 & NA & NA\\
			 Pix3D & 9 & 395 & 10,069 & 64.5 \\
			\hline
		\end{tabular}
		\end{adjustbox}
		\end{center}
	\caption{Datasets with the statistics on the number of classes and samples in each. Standard deviation is calculated for the number of 3D models per main class, sub-classes were not considered.}
\label{tab:dataset_analysis}
\end{table}

The datasets that are used by the majority of the researchers studying geometric deep learning usually consist of a heterogeneous set of items divided into several categories. While being a good fit for research frameworks, these datasets could hardly be used in industry due to the lack of specificity. One major issue that arises while dealing with 3D datasets is class imbalance which introduces a bias in the learning process if not managed properly, as in Table~\ref{tab:dataset_analysis}.

\begin{table}[!ht]
\begin{center}
\begin{adjustbox}{max width=\columnwidth}
        
		\begin{tabular}{|l|c|c|c|c|c|}
		\hline
			\textbf{Dataset} &
			\textbf{Scale} &
			\textbf{LoD} &
			\textbf{Class} &
			\textbf{3D Models} &
		    \textbf{Images}\\
			\hline
			Random3DCity    & 0        &  1-4   &   31   &    NA            &   NA   \\
			3DCityDB (Berlin)& 0        &    2   &   NA   &     550.000      &  NA  \\ 
			Structured3D    & 2        &    4   &   NA   &      NA          &  196K  \\ 
			InteriorNet     & 2        &    4   &   158  &    1M models     &  20M   \\
			SceneNet RGB-D  & 2        &   4    &   37   &      NA          &  5M    \\
			Matterport3D    & 2        &   4    &   20   &    90 scenes     &  194K  \\ 
			Replica         & 2        &   4    &   88   &     18 scenes    &   20M   \\
			Gibson          & 1,2      &   4    &   NA   &     572 scenes   &   NA    \\
			HyperSim        &  2       &   4    &   NA   &     461         &  77.4K  \\
		\hline
	\end{tabular}
	\end{adjustbox}
	\end{center}
\caption{Comparison of architectural datasets acoording to scale, data available, format, number of classes and LoD. Scale is presented as a set of categories \{0, 1, 2\}, where:\\ 0 - city scale, 1 - building scale, 2 - interior scale}
\label{tab:urban_dataset_analysis}

\end{table}

\begin{table}[!ht]
\begin{center}
\begin{adjustbox}{max width=\columnwidth}
	
	\begin{tabular}{|l|c|c|c|c|c|}
	\hline
     \textbf{Dataset} &\textbf{Render}& \textbf{Depth} & \textbf{Seg} & \textbf{Nrm} & \textbf{Format}\\ \hline
	 Random3DCity     &   \checkmark   &                &              &            & CityGML  \\
	 3DCityDB         &  \checkmark    &                &              &            & CityGML  \\
	 Structured3D     & \checkmark     & \checkmark     & \checkmark   &            & Mesh     \\
	 InteriorNet      &   \checkmark   &    \checkmark  &   \checkmark & \checkmark & Mesh       \\
	 SceneNet         &    \checkmark  &                &              &            & RGB-D    \\
	 Matterport3D     &   \checkmark   &   \checkmark   &  \checkmark  &            & RGB-D    \\ 
	 Replica          &    \checkmark  &                &  \checkmark  &            &  Mesh     \\
	 Gibson           & \checkmark     &  \checkmark    &  \checkmark  & \checkmark & Mesh     \\
	 HyperSim         &   \checkmark   &   \checkmark   &  \checkmark  &            & Mesh     \\
	\hline
    \end{tabular}
    \end{adjustbox}
    \end{center}
\caption{Comparison of annotations provided by popular architectural Datasets: Render(RGB), Depth maps, Segmentation masks, Normal maps and 3D model formats. }
\label{tab:urban_dataset_annotation}
\end{table}

As can be observed from the Table ~\ref{tab:urban_dataset_analysis} the datasets that could be used for the application of Deep Learning in the architectural field are skewed to urban scale and interior scale \cite{Matterport3D} \cite{replica} \cite{hypersim} \cite{InteriorNet18} \cite{scenenet} \cite{structured3d} leaving a building scale gap. To our knowledge there is only one open-source dataset\cite{gibson} that provides 3D models of the building structures. This dataset is rather rich in annotations as it provides depth, normals and segmentation annotations, rendered images as well as full 3D mesh models. Unfortunately, this dataset, too, is focused primarily on the interior space while leaving the building exterior without semantic segmentation and texturing. Our dataset concentrates on this limitation by providing a dataset similar in structure but applied to the external appearance of the buildings rather than the internal one. Another drawback of Gibson dataset is its size, as it does not provide a sufficient number of building samples for geometric deep learning frameworks. Our solution for this problem is generation of an arbitrary amount of building samples depending on the need of the research project.
\par At the same time, the unavailability of the datasets or data generation pipelines in the field of architecture results in unavailability of the benchmarks related to the data generation pipeline evaluation.

\subsection{Requirements}\label{requirements}
From the analysis of the existing datasets we have inferred the requirements for the new framework:
\begin{itemize}
\setlength\itemsep{0em}\setlength\parskip{0em}\setlength\topsep{0em}\setlength\partopsep{0em}\setlength\parsep{0em} 
    \item Sufficient dataset size
    \item Class balance to avoid bias
    \item Modularity of data
    \item Variation and diversity in data
    \item Generating required ground truth signals
    \item Associated metadata generation
    \item Extensibility, to accommodate more classes/models
\end{itemize}

\par Since this dataset is mostly intended to be used in Deep Learning tasks, it is necessary for it to have a sufficient amount of samples for the artificial neural networks to learn successfully. The dataset should also be rich in details and alterations of materials, modules, shapes, dimensions. At the same time all the classes and variations should not only be ample in the number of samples but also balanced in order not to introduce an additional bias to the learning process. 
\par The modularity requirement of the dataset that refers to the parametric nature of a generated building allows for the substitution of different parameters and architectural components, randomization, and modification of the object's structure.
\par Architecture is a wide field with multiple problems that can be solved via deep learning. To accustom to as many problems from the spectrum as possible it fundamental to provide different ground truth signals and metadata that would help in the quantitative evaluation of the models' performance. Moreover, the wide range of possible applications requires the dataset to be modular and extendable to be adjusted to different problems based on the researchers' needs.

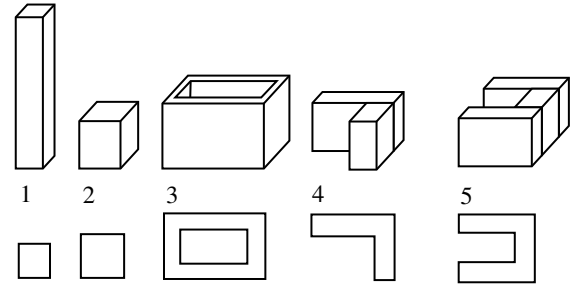
\begin{figure}[!h!]
\centering
\tikzset{every picture/.style={line width=0.75pt}} 

\begin{tikzpicture}[x=0.75pt,y=0.75pt,yscale=-0.55,xscale=.5]

\draw   (12,25.82) -- (23.61,14.21) -- (50.7,14.21) -- (50.7,152.59) -- (39.09,164.2) -- (12,164.2) -- cycle ; \draw   (50.7,14.21) -- (39.09,25.82) -- (12,25.82) ; \draw   (39.09,25.82) -- (39.09,164.2) ;
\draw   (75.42,120.83) -- (93.05,103.2) -- (134.2,103.2) -- (134.2,146.57) -- (116.57,164.2) -- (75.42,164.2) -- cycle ; \draw   (134.2,103.2) -- (116.57,120.83) -- (75.42,120.83) ; \draw   (116.57,120.83) -- (116.57,164.2) ;
\draw   (158.42,104.7) -- (183.92,79.2) -- (285.2,79.2) -- (285.2,138.7) -- (259.7,164.2) -- (158.42,164.2) -- cycle ; \draw   (285.2,79.2) -- (259.7,104.7) -- (158.42,104.7) ; \draw   (259.7,104.7) -- (259.7,164.2) ;
\draw   (186.63,84.2) -- (272.51,84.2) -- (257.07,99.2) -- (171.2,99.2) -- cycle ;
\draw    (186.63,84.2) -- (186.2,99.2) ;
\draw   (308.2,104.2) -- (318.2,94.2) -- (399.2,94.2) -- (399.2,138.17) -- (389.2,148.17) -- (308.2,148.17) -- cycle ; \draw   (399.2,94.2) -- (389.2,104.2) -- (308.2,104.2) ; \draw   (389.2,104.2) -- (389.2,148.17) ;
\draw  [fill={rgb, 255:red, 255; green, 255; blue, 255 }  ,fill opacity=1 ] (345.2,121.2) -- (362.2,104.2) -- (389.2,104.2) -- (389.2,148.2) -- (372.2,165.2) -- (345.2,165.2) -- cycle ; \draw   (389.2,104.2) -- (372.2,121.2) -- (345.2,121.2) ; \draw   (372.2,121.2) -- (372.2,165.2) ;
\draw   (479.2,91.2) -- (489.2,81.2) -- (564.2,81.2) -- (564.2,125.17) -- (554.2,135.17) -- (479.2,135.17) -- cycle ; \draw   (564.2,81.2) -- (554.2,91.2) -- (479.2,91.2) ; \draw   (554.2,91.2) -- (554.2,135.17) ;
\draw  [fill={rgb, 255:red, 255; green, 255; blue, 255 }  ,fill opacity=1 ] (510.2,108.2) -- (527.2,91.2) -- (554.2,91.2) -- (554.2,135.2) -- (537.2,152.2) -- (510.2,152.2) -- cycle ; \draw   (554.2,91.2) -- (537.2,108.2) -- (510.2,108.2) ; \draw   (537.2,108.2) -- (537.2,152.2) ;
\draw  [fill={rgb, 255:red, 255; green, 255; blue, 255 }  ,fill opacity=1 ] (454.2,118.2) -- (464.2,108.2) -- (537.2,108.2) -- (537.2,152.17) -- (527.2,162.17) -- (454.2,162.17) -- cycle ; \draw   (537.2,108.2) -- (527.2,118.2) -- (454.2,118.2) ; \draw   (527.2,118.2) -- (527.2,162.17) ;
\draw   (15,233) -- (46.2,233) -- (46.2,264.2) -- (15,264.2) -- cycle ;
\draw   (77,224.2) -- (120.2,224.2) -- (120.2,264.2) -- (77,264.2) -- cycle ;
\draw   (160,205.2) -- (261.2,205.2) -- (261.2,265.2) -- (160,265.2) -- cycle ;
\draw   (176.2,220) -- (243.2,220) -- (243.2,251.2) -- (176.2,251.2) -- cycle ;
\draw   (390.2,266.2) -- (370.2,266.2) -- (370.2,226.2) -- (307.2,226.2) -- (307.2,206.2) -- (390.2,206.2) -- cycle ;
\draw   (530.2,268.2) -- (454.2,268.2) -- (454.2,249.8) -- (510.2,249.8) -- (510.2,222.9) -- (454.33,222.9) -- (454.33,206.2) -- (530.2,206.2) -- (530.2,268.2) -- cycle ;

\draw (13,177) node [anchor=north west][inner sep=0.75pt]   [align=left] {1};
\draw (76,178) node [anchor=north west][inner sep=0.75pt]   [align=left] {2};
\draw (159,178) node [anchor=north west][inner sep=0.75pt]   [align=left] {3};
\draw (305,177) node [anchor=north west][inner sep=0.75pt]   [align=left] {4};
\draw (453,179) node [anchor=north west][inner sep=0.75pt]   [align=left] {5};
\end{tikzpicture}
\caption{Initial types of the building envelopes. 1 - skyscraper; 2 - isolated building; 3 - patio; 4 - L-shaped building; 5 - C-shaped building.} 
\label{fig:building_types}
\end{figure}

\subsection{Proposed framework}
The synthetic data generation pipeline developed by us addresses the prerequisites mentioned in the Subsection~\ref{requirements}. It is intended primarily for the 3D reconstruction based on parts assembly task \cite{partnet20}, as we have identified that the image to 3D translation would have a major impact in many applications in the architectural field. Although, it should be noted that the annotations and modularity of the framework make it possible to use the generated data for the other deep learning tasks as well.
\par In order to adjust for all the possible applications of this dataset it was decided to partially engage the user in the dataset creation and give the freedom of setting a number of parameters that define the content of the dataset and its variations.
\newline It has been decided to define five classes at the initial stage of the framework development, as can be seen in Figure~\ref{fig:building_types}.

\subsection{Implementation details}
In contrast to the existing generic 3D datasets, and alike many architectural datasets, ours contains synthetic data only. This difference allows generating the data relative to a particular research problem in a customized manner. This characteristic automatically lifts the problem of image-shape alignment present in the other datasets \cite{pix3d} as the images, their segmentation masks, and depth annotations are generated from the mesh directly. For each sample we provide the 3D model in .obj format, the point cloud consisting of 2048 points, the rendered image, the ground truth annotations (segmentation mask, depth image in .exr format, surface normals), and the metadata, which will be expanded in the future editions.

\begin{figure*}[ht!]
    \centering
    \includegraphics[width=1.0\textwidth]{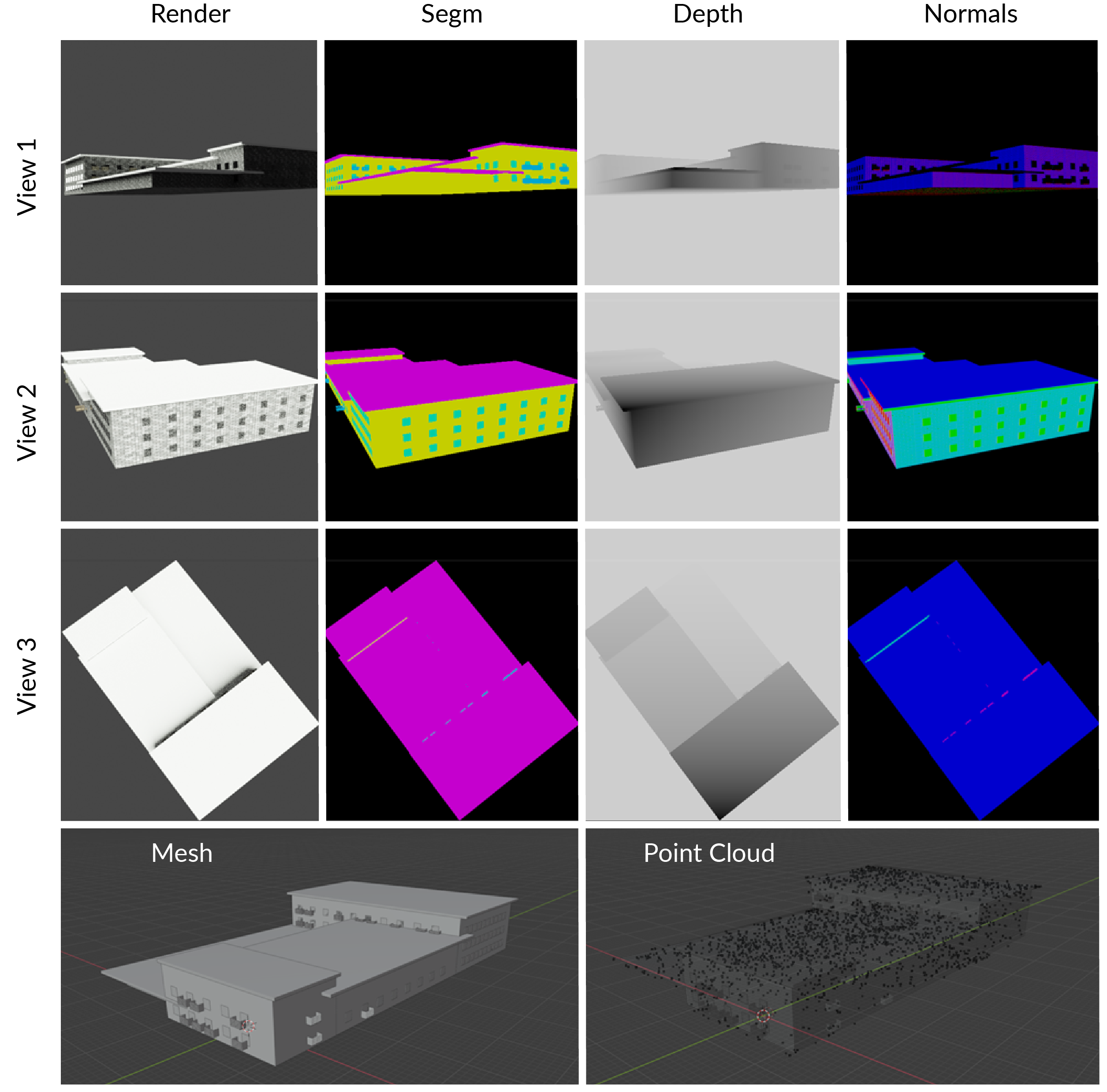}
    \centering
    \caption{Sample of one 3D model generated by the pipeline with its annotations. 3 rows are 3 views of the model, fourth row shows the mesh model and the point cloud generated for the building model. The annotation columns are: 1 - rendered image, 2 - segmentation mask, 3 - depth image, 4 - image of surface normals.}
    \label{fig:samples}
    \centering
\end{figure*}

\subsection{Generation} \label{generation}

The dataset size does not have an upper bound as it is created in a generative manner, the user can define the number of samples needed to be created and the classes the dataset should contain, as well as the set of component modules (e.g. windows and balconies) and textures to include. The user can add his own modules as .obj files to be used in the building dataset generation.
\par 
The input parameters include level of detail (1 or 2), use of materials, textures used (if any), modules used (if any), building types, generated image size, number of points per point cloud, some of the output formats and the general characteristics of the buildings defined by common knowledge and city regulation base, for instance, minimum and maximum building height, width and length. It is also possible to indicate whether it is necessary to generate several views for one 3D building model and how many. All the parameters need to be specified in the configuration file.

\par
The output consists of the following set of data, also illustrated in Figure~\ref{fig:samples}:
\begin{itemize}
\setlength\itemsep{0em}\setlength\parskip{0em}\setlength\topsep{0em}\setlength\partopsep{0em}\setlength\parsep{0em} 
    \item Rendered image of a building, [.png format]
    \item Semantic segmentation mask of a building (each component has its own color) [.png format]
    \item Depth map of a building [.png and .exr format; .png contains normalized values and .exr contains the actual values]
    \item Map of surface normals showing the angle of each surface with respect to the camera [.png format]
    \item 3D model of a building (mesh) [.obj format]
    \item Point cloud of a building, by default point cloud consists of 2048 points which can be changed by the user [.ply format]
\end{itemize}

Dataset Generation code, and sample dataset renders has been made publicly available.
\footnote{\href{https://github.com/CDInstitute/Building-Dataset-Generator}{github.com/CDInstitute/Building-Dataset-Generator}}

\subsection{Domain Randomization}

In order to adjust for the domain adaptation of the real-world building data, it was necessary to introduce domain randomization, which is considered one of the most successful techniques to handle the transfer from synthetic to real data\cite{domain_randomization}. The pipeline uses majorly the parameters and the textures that are relatively close to the real-world ones in order to be able to adjust for the domain adaptation task. Our dataset generation framework has a parameter that allows to create randomized sample images from the same 3D model sample. 
\par The randomization parameters are:

\begin{itemize}
\setlength\itemsep{0em}\setlength\parskip{0em}\setlength\topsep{0em}\setlength\partopsep{0em}\setlength\parsep{0em} 
    \item Texture, reflectance of various building components
    \item Light color, strength and position
    \item Camera position and angle
\end{itemize}

\begin{figure}[h!]
    \centering
    \includegraphics[width=1.0\columnwidth]{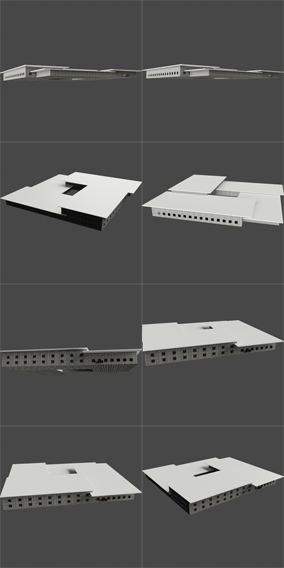}
    \caption{One 3D model sample with multiple views illustrating camera view randomization. The number of possible views per model is defined by the user.}
    \label{fig:camera_randomization}
\end{figure}

\begin{figure}[h!]
    \centering
    \includegraphics[width=1.0\columnwidth]{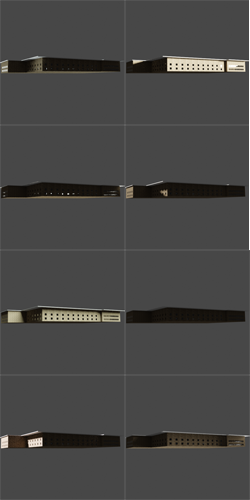}
    \caption{One 3D model sample with multiple views illustrating light randomization. To better illustrate the changes in light the Figure features one view from one model, while in the dataset generation the light changes with the other randomization parameters.}
    \label{fig:light_randomization}
\end{figure}

\begin{figure}[h!]
    \centering
    \includegraphics[width=1.0\columnwidth]{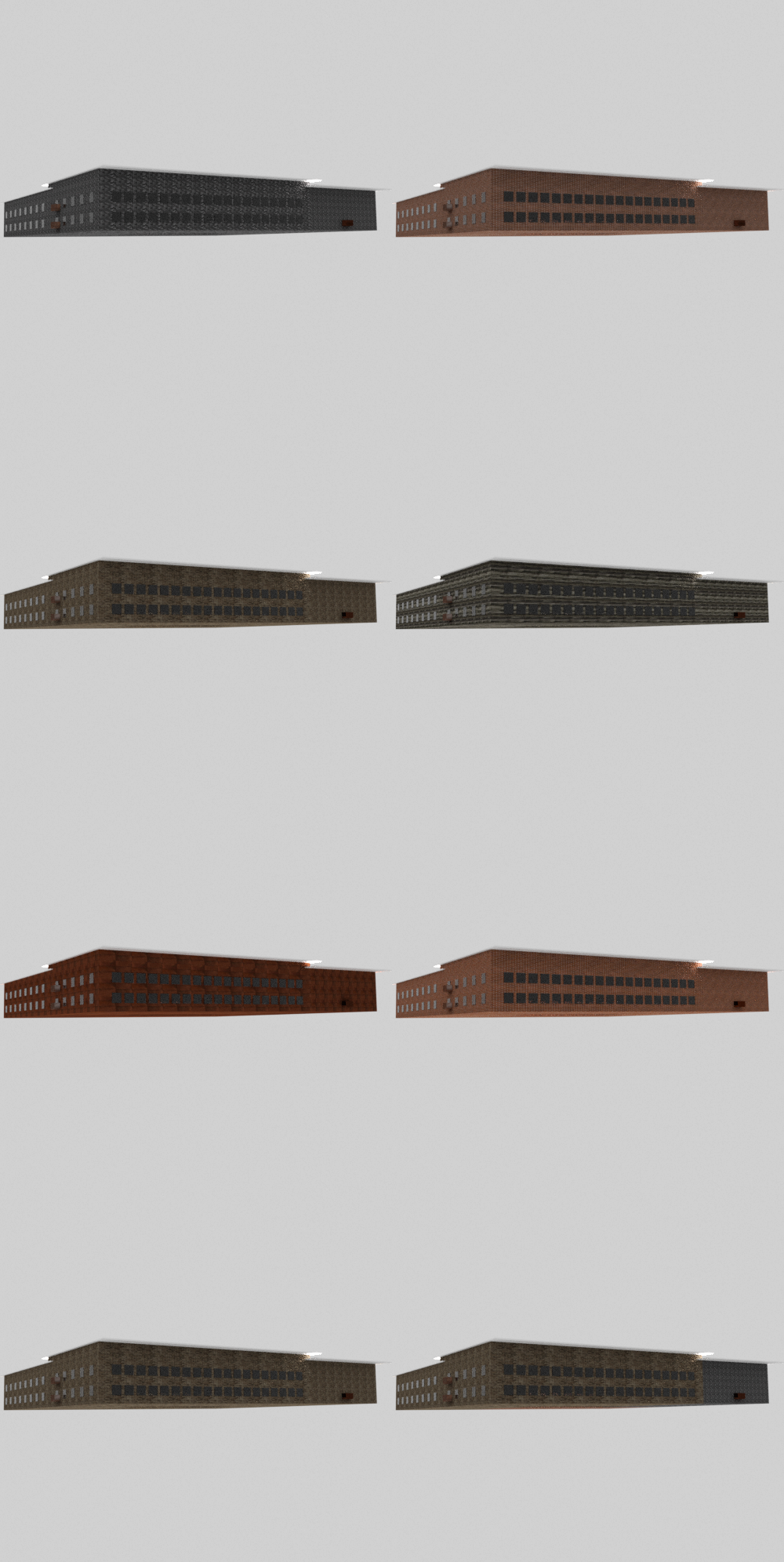}
    \caption{One 3D model sample with multiple views illustrating texture randomization. To better illustrate the changes in texture the Figure features one view from one model, while in the dataset generation volume textures change with the other randomization parameters.}
    \label{fig:texture_randomization}
\end{figure}

Randomization of the camera angle is illustrated in Figure~\ref{fig:camera_randomization}. The camera assumes a random position on the XY plane, while its rotation along the local Z axis is limited by the range (40, 100) in order to take the most significant views of the building model, close to the ones that could be taken in the real world.
Texture is randomized per building volume, as in Figure~\ref{fig:texture_randomization}, with a user-defined probability that all the volumes will be textured with the same material. The textures are selected randomly from the user-added textures.
Light randomization is illustrated in Figure~\ref{fig:light_randomization}.

\subsection{Performance}
We ran the dataset generation algorithm with different input parameters on Windows 10 OS on CPU and GPU using AMD Ryzen 7 3800-X 8-Core Processor and GeForce GTX 1080 and report the performance in the Table~\ref{tab:dataset_performance}. The images were rendered with 500x500 dimensions. The dataset generation time heavily depends on the number of components in the models, as generation of more components requires more time; thus, larger or taller buildings require more time to be generated with respect to the models with smaller dimensions.
\par As previously mentioned, due to the lack of accessible datasets on the architectural scale and the specificity of the contribution, there are no benchmarks and downstream metrics available in the field.

\begin{table}[!ht]
\begin{center}
\begin{adjustbox}{max width=\columnwidth}

	\begin{tabular}{|c|c|c|c|}
	\hline
      \textbf{GPU} & \textbf{EXR}  & \textbf{Multiview} & \textbf{Time (hours)}\\
     \hline
	              &             &             &  1.7     \\
	              &             &  \checkmark &  2.7     \\ 
	              &  \checkmark &             &  1.17    \\
	              &  \checkmark &  \checkmark &  2.5     \\
	   \checkmark &             &             &  0.34    \\
	   \checkmark &             &  \checkmark &  0.8     \\ 
	   \checkmark &  \checkmark &             &  0.92    \\ 
	   \checkmark &  \checkmark &  \checkmark &  0.88    \\ 
	 \hline
    \end{tabular}
    \end{adjustbox}
    \end{center}
	\caption{Comparison of dataset generation performance across different configuration settings(on a Windows 10 OS). The generated dataset consisted of 100 model each, with 3 captured views in multi-view setting.}
\label{tab:dataset_performance}
\end{table}
\section{Limitations}
Building design can be considered a product design task of a very high complexity as it involves multiple parameters densely interconnected between them. This aspect makes the generation of lifelike synthetic data rather complicated. One of the limitations of the proposed dataset generation framework is the lack of authenticity, as it does not provide the real-world data and many visual features present in the photos made in an urban environment lack in the rendered images.
\par Another limitation is the lack of the urban surroundings around the building. The object is rendered in an empty scene which does not occur in the real world and thus makes domain transfer task more complicated. Moreover, the buildings generated with the framework account for the most typical design patterns present in the major cities, while the more creative and outstanding operas of architecture are not included in the generation pipeline due to the high complexity of the problem. This limitation is the most severe one as it narrows the range of possible building solutions and the entire variety of the architectural forms is not presented in this work. Not only does it affect the generalization capability of the neural networks, but it also introduces a bias in the training process.
\par We are planning to address the mentioned limitations in our future research work.
\section{Perspectives}

The proposed dataset is not free from limitations, which we intend to tackle in our future works. This work is the first step towards the use of Geometric Deep Learning in the field of Architecture, consequently, it opens multiple possible ways for future development. One of them is to improve the quality and variation of the building models by adding various building types, components and textures and by making the renders look more realistic. Another possible development direction could be the extension of the level of detail and the addition of the building parameters into the dataset as well as the internal building structure. Moreover, it is important to develop the framework to account for the limitation of not having the urban surroundings. Expanding the framework in order to have the nearby buildings, urban furniture, trees and other elements of urban environment is fundamental for 3D reconstruction in architecture.
\par The exploration of the Geometric Deep Learning Frameworks intended for single image to 3D reconstruction, their performance on the present dataset is equally important to the development of the dataset generation framework, as it is the task the dataset was intended for. This direction of research could bring multiple practical implications related to the fields of architecture, gaming, AR/VR, simulation and others. Moreover, with the variety of annotation this dataset provides, it would be possible to use it in the other Deep Learning tasks as well.
\par Finally, this dataset could serve as a base for the advances in the field of generative architectural design via the exploitation of Generative Adversarial Networks and their ability to learn from the visual features. Moreover, the possibility to generate synthetic data in various ways could facilitate domain adaptation from synthetic to real data.
\section{Contribution}
The contribution of this work consists of two principal parts:
\begin{itemize}

    \item A detailed overview of the existing datasets related to the architectural and urban fields as well as the main geometric deep learning frameworks aimed at solving single image to 3D reconstruction task for buildings;
    \item Providing the architectural and deep learning community with a field-specific dataset generation pipeline targeted at various tasks.
\end{itemize}
\begin{figure*}[ht!]
    \centering
    \includegraphics[width=1.0\textwidth]{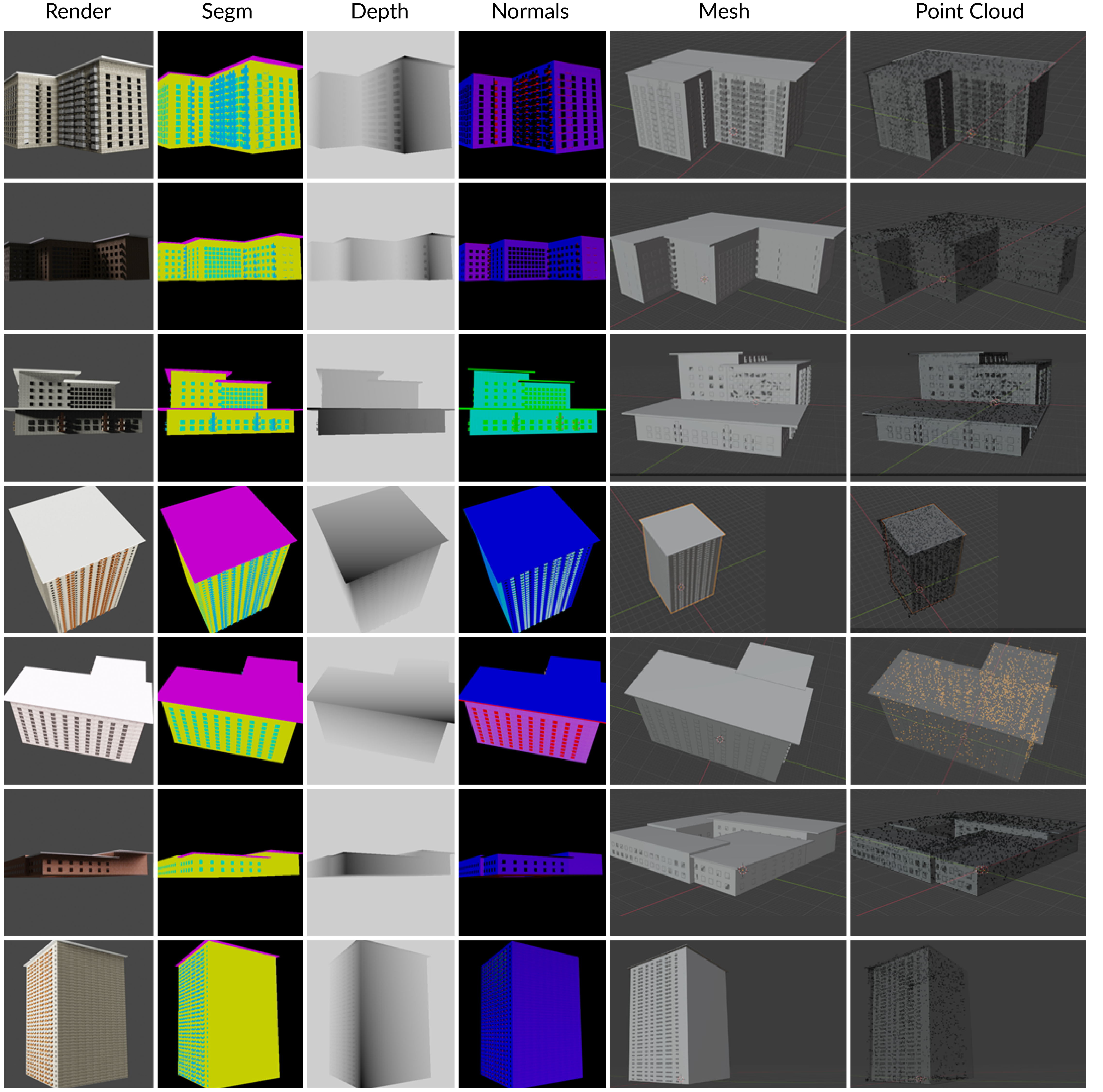}
    \centering
    \caption{Samples of the images generated by the synthetic pipeline. Columns from left to right: rendered images with material textures, semantic segmentation mask, depth map, map of surface normals, image of a mesh object, Point Cloud. Note: point cloud is represented on top of a semitransparent mesh object for the sake of clarity. The point cloud file does not contain any mesh objects.}
    \label{fig:more_samples}
    \centering
\end{figure*}

\section{Conclusion}
\par The overview of the frameworks related to the Single Image to 3D Reconstruction task demonstrate the necessity of a field-specific dataset due to their inability to generalize to unseen data. The exploration of the existing architectures has suggested the necessary requirements and highlighted the weak points of the existing architectural datasets, which later became the foundation for our tool.
\par
Presented dataset generation framework provides a field-specific tool for the generation of 2D and 3D data in various formats. This instrument gives the researchers the possibility to apply the Deep Learning and Geometric Deep Learning techniques to the architecture-related tasks. The framework proved to be able to generate large amounts of various data in a short time compared to the traditional methods previously exploited in the building 3D modelling and the consequent image rendering. Moreover, the way the tool was built allows for extendability and customization of the dataset to the specific tasks within the architectural field using task-specific building components, textures and building typologies. It also addresses the limitations of the existing urban datasets we have given an extensive overview of.
\par However, this dataset is not without drawbacks, the main of which is the use of the synthetic data instead of the real-world data. This is the trade-off that was made due to the high complexity and time and effort consumption of the 3D modelling and 3D scanning processes. Our decision in favor for the synthetic data opens multiple directions for the future work as on the dataset framework as on the algorithms using this data to learn. Moreover, the provided synthetic data facilitates the research in 3D reconstruction related to the architecture field that was not available in the open access previously.


\section{Acknowledgements}
We would like to thank Andrea Giordano, Georgia Gkioxari,  Kaichun Mo, Silvio Savarese, Martin Fischer, Andrea Tagliasacchi, Nicolas Chaulet, Lamberto Ballan, Dmitry Kudinov, Mohammed Keshavari, Andrean Zani, Ignacio Garcia Dorado for valuable discussions and inputs. Furthermore, we are grateful and thankful to the Pytorch, Pytorch3d communities for their help and support. Finally, we would like to acknowledge contributions through helpful discussions, technical support, GPUs donations and Hardware donated by NVIDIA, and the Computational Design Institute.  
{
	\begin{spacing}{1.17}
		\normalsize
		\bibliography{references} 
	\end{spacing}
}

\end{document}